%% file: root.tex
\documentclass[letterpaper, 10 pt, conference]{ieeeconf}  

\IEEEoverridecommandlockouts
\usepackage{graphics}
\usepackage{epsfig} 
\usepackage{mathptmx}
\usepackage{times} 
\usepackage{amsmath}
\usepackage{amssymb} 
\usepackage{multirow} 
\usepackage{todonotes}
\usepackage{xcolor}
\usepackage{bm}

\title{\LARGE \bf Tactile Neural De-rendering}

\author{Jose A. Eyzaguirre \quad Miquel Oller \quad Nima Fazeli
\thanks{Supported by NSF CAREER \#2337870,  NRI \#2220876, and Agencia Nacional de Investigacion y Desarrollo (ANID) of Chile.}
\thanks{Department of Robotics, University of Michigan, USA 
        {\tt\small <jneyza,oller,nfz>@umich.edu}}%
}

\renewcommand\vec{\mathbf}

\begin{document}

\maketitle
\thispagestyle{empty}
\pagestyle{empty}

\begin{abstract}
    Tactile sensing has proven to be an invaluable tool for enhancing robotic perception, particularly in scenarios where visual data is limited or unavailable. However, traditional methods for pose estimation using tactile data often rely on intricate modeling of sensor mechanics or estimation of contact patches, which can be cumbersome and inherently deterministic. In this work, we introduce Tactile Neural De-rendering, a novel approach that leverages a generative model to reconstruct a local 3D representation of an object based solely on its tactile signature. By rendering the object as though perceived by a virtual camera embedded at the fingertip, our method provides a more intuitive and flexible representation of the tactile data. This 3D reconstruction not only facilitates precise pose estimation but also allows for the quantification of uncertainty, providing a robust framework for tactile-based perception in robotics.
\end{abstract}

\input{sections/1-introduction}

\begin{figure}[!t]
  \centering
  \includegraphics[width=\columnwidth]{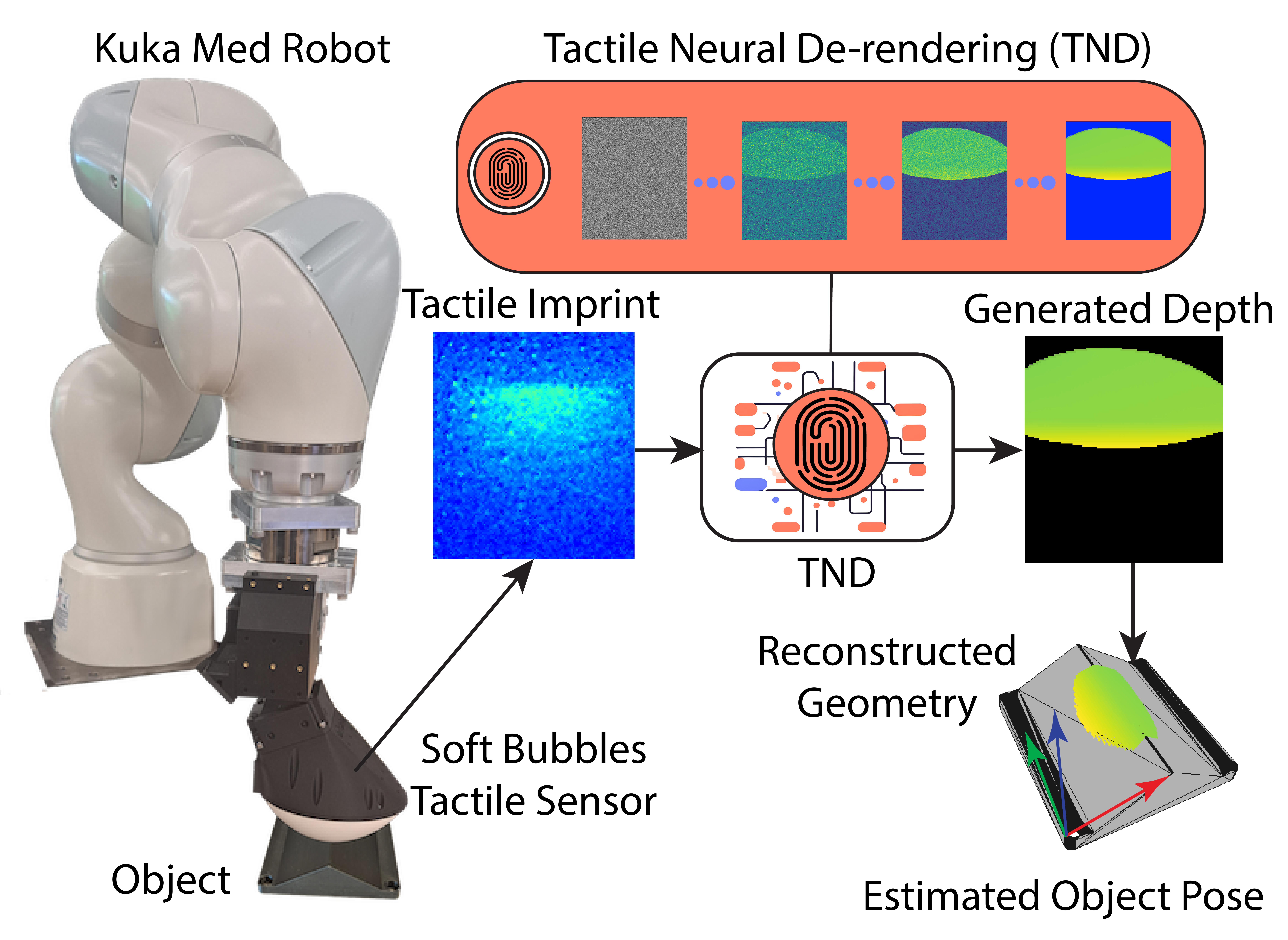}
  \caption{Method overview: A tactile signature is measured from the contact between a visio-tactile sensor and an object, and then De-rendered with our method to produce local 3D in-contact object geometry. The resulting geometry can be used for downstream applications. Here, we focus on pose estimation.}
  \label{fig:teaser}
\end{figure}

\input{sections/2-related-work}

\input{sections/3-problem-statement}

\input{sections/4-method}

\input{sections/5-experiments}

\input{sections/6-discussion}

\addtolength{\textheight}{-6cm}


\bibliographystyle{IEEEtran}
\bibliography{references}

\end{document}

%% file: sections/1-introduction.tex
\section{INTRODUCTION}

Tactile sensing is an essential component of robotic manipulation, particularly in scenarios involving heavy visual occlusion. While vision-based methods excel in open environments, their effectiveness diminishes in cluttered or constrained spaces where objects are partially or entirely obscured from view. Tactile sensors have proven indispensable for filling this perceptual gap, enabling robots to infer vital information about the physical properties of objects through touch. Recent advances in tactile sensing have empowered robots with capabilities such as in-hand pose estimation \cite{bauza_tac2pose, caddeo2023collision, tactile_pose_policy, bubble_patch_pose}, slip detection \cite{dong2019maintaining, james2020slip, sui2021incipient}, force estimation \cite{tactile_fem, daolin_inverse_fem, bubble_fem}, and object recognition \cite{tactile_recognition_early, tactile_obj_recog_humanoid}. These tactile-based capabilities are crucial for executing dexterous manipulation tasks in the presence of uncertainty and occlusion, making tactile perception a key enabler of robust systems.

Despite these advances, existing approaches to tactile perception frequently rely on intricate and domain-specific modeling of sensor mechanics \cite{manipulation_via_membranes, seed, oller2024tactile, wineural} or the estimation of contact patches \cite{daolin_contact_sensing, bubble_patch_pose, bubble_fem}. These methods, while functional, suffer from a number of limitations. First, sensor models often involve complex, high-dimensional representations that are computationally expensive and slow to update in real time. Second, the reliance on intermediate representations, such as contact patches, introduces supervision challenges, as these quantities are difficult to estimate and prone to noise. Third, many tactile perception pipelines are not adaptive to long-term sensor wear and degradation, which can reduce the accuracy of the models over time. Finally, most of these methods are inherently deterministic, providing a single output per tactile input and failing to capture uncertainty in the tactile sensing process.

In this work, we propose Tactile Neural De-rendering, a novel framework that addresses these challenges by drawing inspiration from recent advances in generative modeling. Our approach leverages a generative model to reconstruct a local 3D representation of an object from its tactile signature, akin to rendering the object as though perceived by a virtual camera embedded at the robot’s fingertip. This 3D representation captures the geometry of the object and provides an intuitive, spatially grounded description of the tactile data. By conditioning the generative model on tactile inputs, our method enables a variety of downstream applications including precise pose estimation and the quantification of uncertainty, making it robust to sensor noise and degradation. We demonstrate that this approach not only provides significant improvements in pose estimation accuracy but also offers a more flexible, adaptive framework for tactile-based perception in robotic manipulation.



%% file: sections/2-related-work.tex
\section{RELATED WORK}

\subsection{Vision-Based Tactile Contact Estimation}
Vision-based tactile sensors such as the GelSight \cite{gelsight}, GelSlim \cite{gelslim3}, DIGIT \cite{digit}, FingerVision \cite{fingervision}, and TacTip \cite{tactip} perceive contact interactions via observing the deformations of an elastic material in contact with the environment, typically an elastomer. These sensors provide high-resolution information of the contact interaction between the sensor and the environment. However, these observations tend to suffer from artifacts produced by the elastomer, such as tactile shadowing. This is especially the case for highly deformable tactile sensors such as the Soft Bubbles \cite{soft_bubbles} since the highly compliant membrane deforms continuously producing deformation beyond the actual contact. 
As a result, some techniques have been proposed to estimate the contact patches from tactile observations. For instance, \cite{bubble_patch_pose} model the sensor mechanics and recover the contact patches by solving a QP optimization problem. Similarly, \cite{bubble_fem} present a finite element model (FEM) of the sensor mechanics that simultaneously estimates the force response and the contact patches. Last, \cite{wineural} use Physics Informed Neural Networks (PINNs) to model such deformations and estimate the contact patches. However, these methods require domain-specific modeling of the sensor mechanics and expensive computations. As opposed to these methods, our approach estimates the contact patches by learning to reconstruct the in-contact object's geometry from tactile observations.

\subsection{Tactile 3D Shape Reconstruction}
Several methods have emerged to generate 3D representations from tactile data. For instance, \cite{touchsdf} leverage deep learning techniques like DeepSDF to map tactile images to local surface meshes and predict signed distance functions (SDF) for reconstructing continuous 3D shapes from touch. 
Similarly, \cite{meshes_from_tactile} iteratively refines a mesh based on tactile interactions. 
Furthermore, \cite{active3Dreconstruction} employs high-resolution vision-based tactile sensors and data-driven models to guide shape exploration, enabling more efficient tactile object understanding. 
Neural Radiance Fields (NeRF) have also been adapted to fuse tactile and visual data by conditioning NeRF models on tactile latent codes to render novel object views \cite{tactile_radiance_fields}. 
Finally, \cite{inhand_reconstruction} proposes an in-hand object reconstruction framework that works for both rigid and deformable objects. 
However, despite the effectiveness of these approaches, none of them utilize diffusion models for generating object geometry from tactile data.

\subsection{Tactile Probabilistic Diffusion}

Diffusion models have emerged as a powerful tool for generative conditional modeling across various modalities, including tactile data. For example, \cite{yang2023generating} utilize diffusion models to generate visual-tactile scenes from touch. 
Similarly, \cite{touch2touch} employ diffusion models for cross-model tactile generation, estimating tactile signals captured from different sensors.
Close to our work, \cite{tactile_diffusion} use probabilistic diffusion models to bridge the sim-to-real gap for the DIGIT \cite{digit} tactile sensors, rendering real-world tactile observations from simulated depth maps. 
However, their focus is on generating tactile observations from object geometry, whereas our approach addresses the inverse problem: generating the in-contact geometry that produces those tactile observations (de-rendering). This inverse process is particularly valuable for tasks such as pose estimation and object identification.

%% file: sections/3-problem-statement.tex
\section{PROBLEM STATEMENT}

In this work, we aim to estimate the geometry of a rigid object in contact with a vision-based tactile sensor. The input to the problem is a tactile sensor image $\vec{y}$, which encodes the deformation of a soft elastomer layer in response to contact with the object. The goal is to recover local 3D geometry $\hat{\vec{x}}$  of the object that best explains the tactile signature, minimizing the difference between the estimated geometry $\hat{\vec{x}}$ and the true but unknown geometry $\vec{x}$. The objective of the downstream pose estimation task is to estimate $\vec{q}\in \mathrm{SE}(2)$ given the estimated object geometry $\hat{\vec{x}}$. We note that the tactile signature is influenced by factors including the interaction between the object’s geometry and the sensor’s physical properties.

We assume the object in contact is rigid, meaning that any deformation occurs exclusively due to that of the sensor, rather than the object itself. This assumption simplifies the relationship between the tactile signature and the object's geometry, as the deformation is fully captured by the sensor's elastic properties.

%% file: sections/4-method.tex
\section{METHOD}

\begin{figure*}[!ht]
  \centering
  \includegraphics[width=\textwidth]{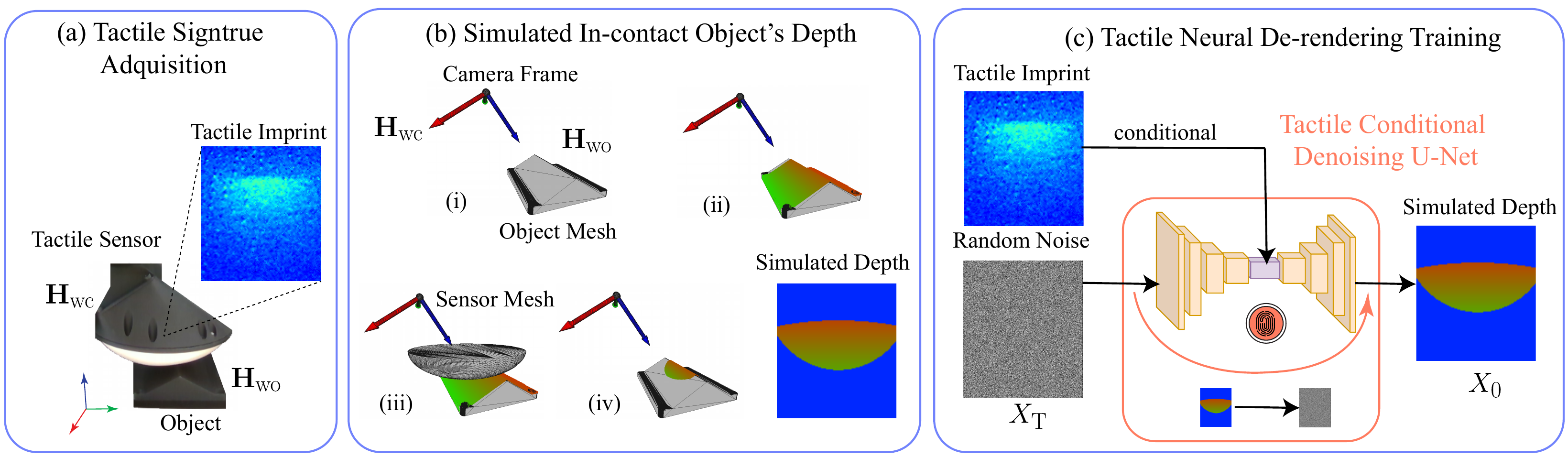}
  \caption{A summary of Tactile Neural De-rendering training. (a) A tactile signature is obtained from the contact between a visio-tactile sensor and an object, with known poses. (b) The four key steps to simulate the in-contact object's depth. (c) Tactile Neural De-rendering training process, we train a diffusion model conditioned on the tactile imprint to generate the in-contact object's depth. For training, we make use of the simulated in-contact object's depth computed in (b) as $X_0$.}
  \label{fig:method}
\end{figure*}

Our approach is composed of three components: (i) \textbf{Simulated in-contact object depth computation}, which takes the sensor parameters as input and outputs a rendered depth map of the simulated in-contact object; (ii) \textbf{Tactile Neural De-rendering}, a process that generates the in-contact object’s depth conditioned on the tactile signature $\vec{y}$; and (iii) \textbf{3D reconstruction of the object geometry}, where the estimated in-contact object geometry $\hat{\vec{x}}$ is recovered from the generated depth map. Fig. \ref{fig:method} illustrates the training process of Tactile Neural De-rendering, which integrates these components.

\subsection{Simulated in-contact object's depth computation}
\label{subsec:sim_obj_depth}

Acquiring real-world data for contact patch estimation is both labor-intensive and prone to error, as it typically requires manual annotation of contact regions, a process that is often subjective and inconsistent. Furthermore, gathering sufficient real-world data is expensive and time-consuming, particularly when the goal is to capture a wide variety of object shapes and interactions with a tactile sensor. This makes it challenging to build large, diverse datasets necessary for training robust models.  

Current simulation frameworks for tactile sensing are extremely limited due to the complexity of modeling the deformations induced by contact between objects and soft materials. Simulating these deformations with high fidelity requires capturing intricate physical phenomena, such as material elasticity and friction, which is computationally expensive and difficult to calibrate. Moreover, there is often a significant mismatch between simulated and real-world data, raising concerns about whether high-fidelity simulation is even beneficial for model training.

Given these challenges, we propose a simpler simulation environment, relying on existing 3D rendering libraries such as Open3D \cite{open3d}. By leveraging simulation, we can generate a wide variety of tactile interactions efficiently and with precise ground truth information, ensuring better model performance and reducing the reliance on costly and error-prone real-world data collection.

Our proposed simulation process consists of four key steps: object depth rendering, depth projection, signed distance field computation, and in-contact depth estimation.

\begin{itemize}
    \item \textbf{Object depth rendering:} To simulate the interaction between a rigid object and the tactile sensor, we first position the object’s mesh $\mathcal{M}_o$ and the virtual depth camera within the simulation environment. The virtual camera is aligned with the sensor’s view as it would appear in real-world experiments. A depth image of the object, denoted $D$, is then rendered. This depth image represents the object from the perspective of the virtual camera embedded at the fingertip of the tactile sensor, without simulating any deformations.
    \item \textbf{Depth projection to 3D:} Once the depth image $D$ is rendered, we project the depth pixels into 3D world coordinates to obtain a set of 3D points, $p_w$. These points represent the surface of the object as seen from the sensor's perspective. The projection is performed using the sensor’s pose relative to the world frame.
    \item \textbf{Signed distance field computation:} After projecting the object’s surface points, we introduce a mesh $\mathcal{M}_e$ representing the sensor physical substrate (e.g., elastomer) in contact. Next, we calculate the signed distance field (SDF) between the substrate's nominal geometry and the projected object points $p_w$. The SDF measures the distance from each point to the substrate, allowing us to determine whether the points lie inside or outside the substrate. We emphasize that we do not model any deformation and instead compute distances between rigid bodies that may be in penetration.
    \item \textbf{In-contact depth estimation:} To simulate the in-contact depth image, we generate a binary mask $M$ by selecting points from $p_w$ that are inside the substrate, as determined by the SDF. Points that lie inside the substrate are considered in contact with the sensor. The final simulated in-contact depth image is obtained by pixel-wise multiplying the object’s original depth image $D$ with the contact mask $M$, yielding $M \odot D$. This multiplication effectively removes points that are not in contact, producing a map that represents the in-contact object's depth as though perceived by a camera embedded at the fingertip of the tactile sensor.
\end{itemize}

\subsection{Tactile Neural De-rendering}

The objective of Tactile Neural De-rendering is to extract features from a tactile signature to generate a plausible geometry that best explains the measurement. To achieve this, we employ Denoising Probabilistic Diffusion Models as image-to-image translators to generate in-contact object depth images.

We build upon the work of \cite{ho2020denoising} and model Tactile Neural De-rendering as a conditional denoising process. Here, $X_0$ represents the in-contact object depth image we aim to generate given a tactile signature $Y$. In the forward process, Gaussian noise $\epsilon$ is added to $X_0$ according to a variance scheduler $\beta_t$. This process is repeated $T$ times until the image is completely degraded into pure isotropic Gaussian noise. For a given timestep $t$, the corresponding latent $X_t$ can be computed as:

\[
X_t = \sqrt{\bar{\alpha_t}}X_0 + \sqrt{1 - \bar{\alpha_t}}\epsilon, \quad \text{where }\alpha_t = 1 - \beta_t \quad \bar{\alpha_t} =  \prod_{s=1}^{t}\alpha_s
\]

Since the forward process permits sampling at a specific $t$, it allows for loss computation at distinct timesteps, thereby enabling efficient training of the model without the need to traverse the entire denoising process. In the reverse process, a network $\epsilon_{\theta}(X_t, Y, t)$ is trained to predict the added noise at each timestep $t$. Our loss function is expressed as follows:

\[
L(\theta) = \mathbb{E}_{t, X_0, \epsilon}[\|\epsilon - \epsilon_{\theta}(X_t, Y, t)\|_{2}^2]
\]

In order to recover $X_0$, we sample the model for each timestep sequentially as follows
\[
X_{t-1} = \frac{1}{\alpha_t}(X_t - \frac{\beta_t}{\sqrt{1-\bar{\alpha_t}}}\epsilon_{\theta}(X_t, t)) + \sqrt{\beta_t}z
\]

Where $z \sim \mathcal{N}(\boldsymbol{0}, \boldsymbol{I})$.

\subsection{3D reconstruction of the in-contact object geometry}

To reconstruct the in-contact object's geometry from the depth image generated by Tactile Neural De-rendering, we utilize the pinhole camera model to project depth data into 3D coordinates using the camera's intrinsic parameters. 


%% file: sections/5-experiments.tex
\section{EXPERIMENTS}

For this study, we evaluate our approach using the Soft-Bubble Sensor v1 \cite{soft_bubbles}. These sensors produce tactile images
based on the interaction between the object’s geometry,
the sensor’s physical substrate, and the sensor’s camera
parameters. The key factors influencing the tactile readings include the elasticity and thickness of the elastomer, the internal air pressure, and the relative pose of the elastomer to the camera, denoted as $H_{cb}$. Additionally, the camera’s internal calibration plays a significant role in shaping the tactile images. Extrinsic parameters, including the object and depth camera poses in the world coordinate frame, $H_{wo}$ and $H_{wc}$, further affect the sensor’s output.  Our goal is to reliably infer the 3D geometry of the object in-contact with the sensor, assuming rigid body interaction.

We demonstrate the effectiveness of Tactile Neural De-rendering through two evaluations: (i) analyzing the quality of the generated in-contact object depth images, and (ii) using the 3D reconstructions for pose estimation. For the first, we compare the generated depth images with corresponding simulated data using distance and image similarity metrics. For the second, we compare our method to a thresholding-based pose estimation approach, and we quantify uncertainty by analyzing the pose estimation error.

The following sections outline the experimental procedures, including dataset generation, model training parameters, and the two evaluation tests that showcase the capabilities of our framework.

\subsection{Dataset generation}

\begin{figure}[!t]
  \centering
  \includegraphics[width=\columnwidth]{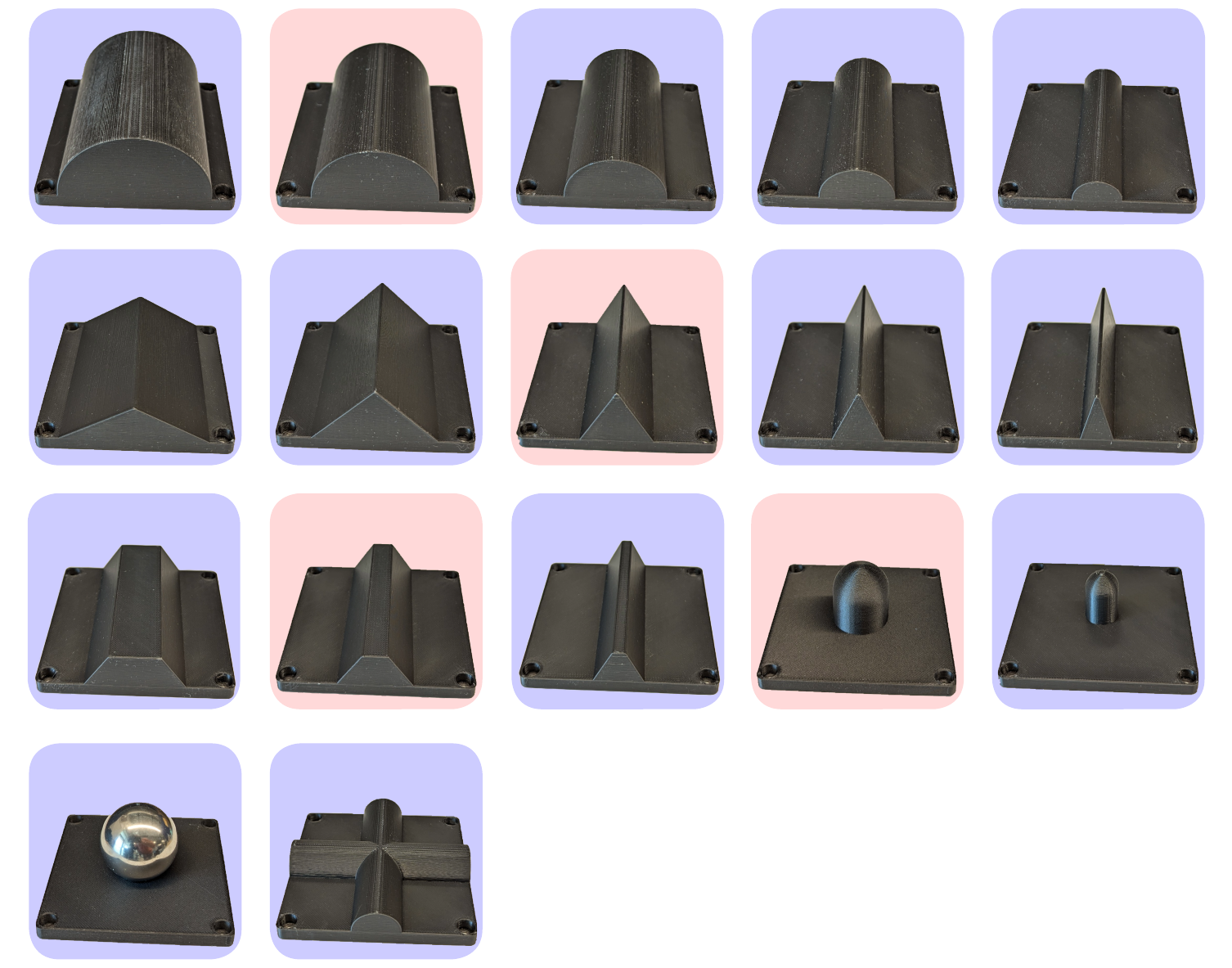}
  \caption{Designed objects to train and test our method. Among them are different sizes of cylinders, triangles, trapezoids, bullets, spheres, and crosses. The objects used for training have a light-blue background, and the ones used for testing unseen objects have a light-red background.}
  \label{fig:objects}
\end{figure}

\begin{table*}[ht]
    \caption{Depth image generation evaluation for different data configurations}
    \label{tab:depth_gen}
    \centering
    \begin{tabular}{|l|cc|cc|cc|cc|cc|cc|}
        \hline
        \multirow{3}{*}{Data Config} & \multicolumn{6}{c|}{Seen Objects - Unseen Poses} & \multicolumn{6}{c|}{Unseen Objects} \\
        \cline{2-13}
        & \multicolumn{2}{c|}{L1 [mm]} & \multicolumn{2}{c|}{PSNR} & \multicolumn{2}{c|}{SSIM} & \multicolumn{2}{c|}{L1 [mm]} & \multicolumn{2}{c|}{PSNR} & \multicolumn{2}{c|}{SSIM} \\
        \cline{2-13}
        & Mean ↓ & Std ↓ & Mean ↑ & Std ↓ & Mean ↑ & Std ↓ & Mean ↓ & Std ↓ & Mean ↑ & Std ↓ & Mean ↑ & Std ↓ \\
        \hline
        Real data & $4.17$ & $2.15$ & 42.50 & 7.09 & 0.84 & 0.049 & $3.12$ & $1.61$ & 44.12 & 7.52 & 0.86 & 0.040 \\
        + synthetic 6k & $3.54$ & $1.79$ & 43.83 & 7.88 & 0.86 & 0.041 & $3.11 $ & $1.79 $ & 45.23 & 8.64 & 0.87 & 0.042 \\
        + synthetic 12k & $2.95$ & $1.57 $ & 44.52 & 7.32 & 0.87 & 0.038 & $2.74$ & $1.66$ & 45.66 & 8.69 & 0.89 & 0.039 \\
        + synthetic 18k & $3.31$ & $1.90$ & 44.33 & 8.20 & 0.86 & 0.048 & $2.83$ & $1.63$ & 45.75 & 8.68 & 0.88 & 0.039 \\
        + synthetic 24k & $3.05$ & $1.55$ & 44.73 & 8.40 & 0.88 & 0.040 & $2.95$ & $1.68$ & 46.64 & 9.71 & 0.89 & 0.039 \\
        \hline
    \end{tabular}
\end{table*}

\begin{table*}[htp]
    \centering
    \caption{Pose estimation evaluation}
    \label{tab:pose_est}
    \begin{tabular}{|l|cc|cc|cc|}
        \hline
        \multirow{2}{*}{Models} 
        & \multicolumn{2}{c|}{Position Error [mm]} & \multicolumn{2}{c|}{$\theta$ Error [rad]} &
        \multicolumn{2}{c|}{Chamfer distance [$m^2$]} \\
        \cline{2-7}
        & Mean ↓ & Std ↓ & Mean ↓ & Std ↓ & Mean ↓ & Std ↓ \\
        \hline
        \textbf{Unseen poses} &  &  &  &  &  & \\
        Baseline & $4.54$ & $16.8$ & $2.7\cdot 10^{-1}$ & $1.14\cdot 10^{0}$ & $5.54\cdot 10^{-5}$ & $3.49\cdot 10^{-4}$ \\
        Real data & $2.19 $ & $12.0$ & $6.43 \cdot 10^{-2}$ & $4.63\cdot 10^{-1}$ & $3.11 \cdot 10^{-5}$ & $3.43 \cdot 10^{-4}$ \\
        + syn. 6k & $1.59 $ & $7.65 $ & $5.27 \cdot 10^{-2}$ & $3.08 \cdot 10^{-1}$ & $1.93 \cdot 10^{-5}$ & $1.92 \cdot 10^{-4}$ \\
        + syn. 24k & $\boldsymbol{1.13 }$ & $\boldsymbol{4.08 }$ & $\boldsymbol{4.26 \cdot 10^{-2}}$ & $\boldsymbol{1.53 \cdot 10^{-1}}$ & $\boldsymbol{1.69\cdot 10^{-5}}$ & $\boldsymbol{9.88 \cdot 10^{-5}}$ \\
        
        \hline
        \textbf{Unseen objects} &  &  &  &  &  &  \\
        Baseline & $5.08$ & $12.8$ & $3.07\cdot 10^{-1}$ & $7.90\cdot 10^{-1}$ & $7.25\cdot 10^{-5}$ & $2.47\cdot 10^{-4}$ \\
        Real data & $2.10 $ & $4.58 $ & $6.08 \cdot 10^{-2}$ & $2.30 \cdot 10^{-1}$ & $2.98\cdot 10^{-5}$ & $7.95 \cdot 10^{-5}$ \\
        + syn. 6k & $1.42 $ & $4.27 $ & $4.29 \cdot 10^{-2}$ & $1.79 \cdot 10^{-1}$ & $2.52 \cdot 10^{-5}$ & $8.75 \cdot 10^{-5}$ \\
        + syn. 24k & $\boldsymbol{1.26}$ & $\boldsymbol{3.77}$ & $\boldsymbol{3.86 \cdot 10^{-2}}$ & $\boldsymbol{1.63\cdot 10^{-1}}$ & $\boldsymbol{2.18\cdot 10^{-5}}$ & $\boldsymbol{7.58 \cdot 10^{-5}}$ \\
        \hline
    \end{tabular}
\end{table*}

We created two datasets to train and test Tactile Neural De-rendering. A synthetic one to pre-train our method, allowing us to create a huge number of depth image pairs in a short period of time, and a real-data one, to finetune our method with real Soft Bubble sensor tactile signatures.

For the real-data dataset we designed and 3D printed 17 objects as can be seen in figure \ref{fig:objects}. Each object has a base that allows us to easily attach it to a table with a grid with known positions, allowing us to control the object pose $H_{wo}$. We attach the bubble sensor to a Kuka LBR iiwa R820, and automate the process of measuring tactile signatures from the interaction between the object and sensor over a wide distribution of relative poses. Given that we know the pose of both the sensor (from arm kinematics) and object, we render the simulated in-contact object's depth as described in section \ref{subsec:sim_obj_depth}. Thus, we obtain depth image pairs from the sensor measurement and the simulated in-contact object's depth. In total, the real dataset has 3.760 depth image pairs.

The synthetic dataset is composed of 24 objects, 14 mesh models of the real dataset, and 10 custom designed objects to increase the diversity of tactile features. To create tactile signatures, we follow a similar procedure as the one outlined in Sec. IV. A) ensuring wide coverage of objects pose with respect to the sensor. For each sensor measurement, we position the objects' mesh at the origin, and a mesh representation of the elastomer’s surface, with the same thickness as the Soft Bubble sensor, at random positions in-contact with the object. Then, we render depth and add tactile shadows by linearly connecting the rendered points from the top of the object to the points on the elastomer at a distance of 21 pixels. Doing so we create a cloak around the object that approximates the artifacts created by the deformation of the elastomer. We added Gaussian noise to account for the measurement noise, with $\mu=0$ and $\sigma=1.38\cdot 10^{-3}$, values estimated from a real sensor's measurements while not in contact. 
In total, the synthetic dataset has 24.000 depth image pairs.

We split the real-data dataset into three: seen objects and poses, seen objects in unseen poses, and unseen objects, the former to train our models, and the latter two to test them. Using 13 out of the 17 objects, we created the seen dataset, from which we divided it into seen and unseen poses using a ratio of 9:1, respectively. For the unseen objects, we use the remaining four objects with all their poses.

\subsection{Tactile Neural De-rendering model and training details}

In order to recover the 3D geometry $\hat{\vec{x}}$ of the object that best explains the tactile signature $\vec{y}$, we use denoising diffusion probabilistic models to generate in-contact object's depth images conditioned on the sensor's imprint. Where the sensor's imprint is defined as the subtraction between a tactile signature while the sensor is not in contact (reference), and the measurement from the sensor. The architecture of the model we use is based on Ho's work \cite{ho2020denoising}, we specifically use their U-Net network as the backbone for the reverse process. We modified the number of input channels to one and conditioned the model concatenating the tactile imprint to each latent. We use 250 denoising steps, and a cosine noise scheduler as presented in \cite{nichol2021improved}.

We trained our model with five different data configurations. The first was trained only with the seen objects and poses dataset, and the four others were pre-trained with subsets of the synthetic dataset of 6k, 12k, 18k, and 24k depth image pairs, and later fine-tuned with the seen objects and poses dataset. Each of these models was trained using L1 loss between the predicted and actual noise, and the Adam optimizer with a learning rate of $1 \cdot 10^{-4}$. We pre-trained the models with the different subsets of the synthetic dataset for 120 to 160 epochs and fine-tuned them for 40 epochs. For the model only trained with the seen objects and poses dataset, we trained it for 160 epochs.

\subsection{Quality of the generated in-contact object's depth images}


\begin{figure}[!t]
  \centering
  \includegraphics[width=\columnwidth]{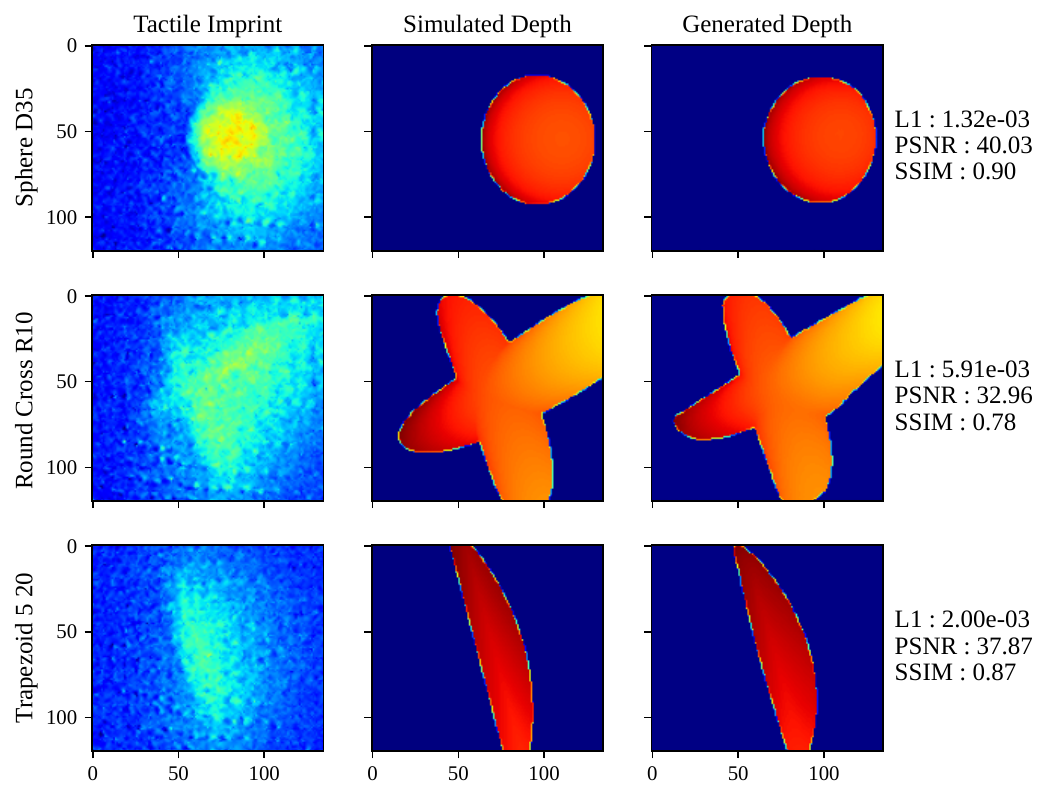}
  \caption{Depth generations examples for the model with data configuration of real-data + synthetic 24k. Each row represents a single object, and each column displays the tactile imprint, the simulated depth used to test the model, and the generated in-contact object depth image, respectively. Each depth image is colored using the JET colormap.}
  \label{fig:generations}
\end{figure}

We evaluate the various models by comparing their generations against the simulated in-contact object's depth. To do so, we use L1 as a distance metric, and PSNR and SSIM as image metrics. We test each model on two scenarios: seen objects in unseen poses, and unseen objects. 

Results are shown in Tab.~\ref{tab:depth_gen} for each metric, indicating that the model is learning useful features from the synthetic datasets, as each metric performs better as it is pre-trained with a larger synthetic dataset. The mean L1 error is on the order of millimeters, and its standard deviation is on the same order of magnitude. In Fig. \ref{fig:generations} we show features examples of the depth generations for the model trained with real data, and synthetic 24k. We display the process from the tactile imprint used to condition the model and the simulated and generated depths.

\subsection{Pose estimation}

The objective of the downstream pose estimation task is to estimate $\vec{q}\in \mathrm{SE}(2)$ given the estimated object geometry $\hat{\vec{x}}$. To test our method we used the models that were trained with the seen objects and poses dataset, and the ones pre-trained with 6k and 24k synthetic image pairs. As a baseline, we used a thresholding-based approach to extract the in-contact section of the sensor's tactile signature. This method takes as input a tactile signature, and the sensor's reference, and returns the section of the tactile signature that diverges in more than a given threshold from the sensor's reference. We set this threshold to 3 mm. 

We project each generated in-contact object's depth images into a point cloud and register it using CHSEL \cite{chsel}. We run five iterations for each of the ten randomly initialized object's mesh poses and keep the one with the best fit.

To quantify the performance of our estimated poses, we measure the Euclidean distance between the estimated and ground truth positions, rotations, and the unidirectional Chamfer distance between two identical point clouds, extracted from the objects' meshes, and transformed using the estimated and ground truth poses. Equation \ref{eq:CD} shows how the unidirectional Chamfer distance is computed.

\begin{equation}
    CD(P_1, P_2) = \frac{1}{|P_1|} \sum_{x \in P_1} \min_{y \in P_2} \|x - y\|_2^2
    \label{eq:CD}
\end{equation}

For positional, and rotational errors, we excluded the object's variables with inherent geometric ambiguities. Several objects in our dataset possess translational ambiguity when sensed along one of their principal axes, such as cylinders, triangles, and trapezoids. Additionally, objects like spheres and bullets exhibit rotational ambiguity, so we do not use them to compute the mean rotational error. 

Table \ref{tab:pose_est} presents the results of our experiment, illustrating that models pre-trained with synthetic data significantly outperformed the baseline. These models demonstrated superior accuracy and robustness across the three evaluated metrics.

We also use the pose estimation error to quantify the uncertainty of the generated in-contact object's geometries. To do so, we generate 50 in-contact object's depth images for a given tactile signature. We register the projected point clouds using CHSEL, but this time we only use one fixed initialization for the source mesh. 

\begin{figure}[!t]
  \centering
  \includegraphics[width=\columnwidth]{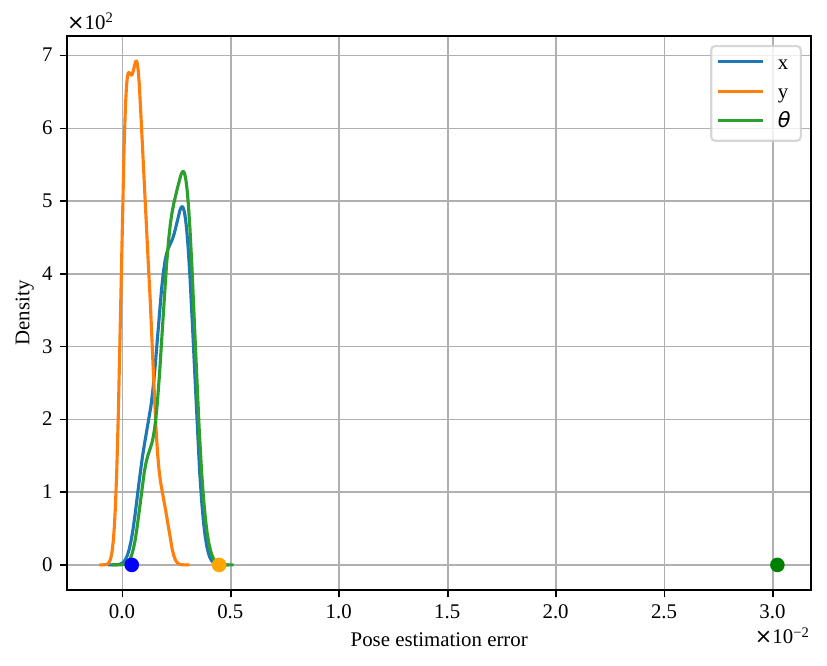}
  \caption{Kernel density estimate (KDE) plot of the pose estimation error using our method, for a specific contact configuration of the Soft Bubble sensor and the Round Cross object. Dots represent the pose estimation error from the baseline as they are deterministic.}
  \label{fig:box_plot}
\end{figure}

In figure \ref{fig:box_plot} we show a kernel density estimate (KDE) plot of the pose estimation error for a specific configuration of the Round Cross object, using the model trained with real-data + synthetic 24k. We also illustrate in the figure dots that represent the pose estimation error from the baseline given the same tactile signature. Our method not only surpasses the baseline in performance but also can quantify the uncertainty of downstream tasks, a capability that the inherently deterministic baseline lacks.

%% file: sections/6-discussion.tex
\section{DISCUSSION}

We have introduced a framework that enhances the understanding of contact geometries captured by a visio-tactile sensor through the use of generative models. This approach allows us to bypass the complexities of modeling the sensor's elastomer mechanics, relying solely on tactile signatures to reconstruct the object's local 3D geometry. Notably, our method was trained exclusively with real tactile data and a simple simulation environment that uses rendering to approximate the ground truth of the in-contact object's depth.

One key limitation of our method is its dependence on precise calibrations of the robotic arm and object attachments during training. The accuracy of the simulated in-contact object's depth relies on known parameters $H_{wo}$ and $H_{wc}$. Calibration errors can introduce biases in the simulated object's depth, which may propagate through the 3D reconstructions, ultimately degrading the method's performance. A similar issue arises if the elastomer mesh is inaccurately modeled—if it is too inflated, the model may hallucinate object features, and if it is too deflated, crucial object details could be lost.

Our experiments highlight four key findings: (i) our method successfully generates in-contact object depth images exclusively from tactile signatures; (ii) the synthetic dataset we developed possesses valuable features for the model, facilitating scalable tactile data creation for pre-training; (iii) our method is well-suited for real-world downstream tasks, such as pose estimation; and (iv) the method can quantify the level of uncertainty for a real-world downstream tasks.

It would be interesting to explore whether improved results could be obtained by utilizing a more diverse dataset than the one used in this study, particularly by incorporating common household objects for testing. Furthermore, evaluating our method with other visio-tactile sensors could provide valuable insights and demonstrate that its effectiveness is not restricted to the Soft Bubble sensor.